\pdfoutput=1
\documentclass{article}

\PassOptionsToPackage{numbers, compress}{natbib}

\usepackage[preprint]{nips_2018}

\usepackage[utf8]{inputenc} 
\usepackage[T1]{fontenc}    
\usepackage{hyperref}       
\usepackage{url}            
\usepackage{booktabs}       
\usepackage{amsfonts}       
\usepackage{nicefrac}       
\usepackage{microtype}      
\usepackage{algorithm}
\usepackage[noend]{algpseudocode}
\usepackage{multirow}

\usepackage{color}
\usepackage{graphicx}

\newtheorem{theorem}{Theorem}

\newcommand{\cH}{\mathcal{W}}

\newcommand{\err}{\mbox{err}}
\newcommand{\errub}{\widehat{\mbox{err}}}
\newcommand{\edge}{\gamma}
\newcommand{\edgeEmp}{\hat{\edge}}
\newcommand{\sign}{\mbox{sign}}

\newcommand{\Dist}{{\cal D}}

\newcommand{\training}{{\cal S}}
\newcommand{\Z}{Z_{\training}}
\newcommand{\Ztest}{{\mathbf Z}}
\newcommand{\neff}{{n_{\mbox{eff}}}}

\newcommand{\tmsn}{{\bf TMSN}}
\newcommand{\Sparrow}{{\bf Sparrow}}

\newcommand{\weakRules}{{\cal W}}
\newcommand{\strongRules}{{\cal H}}

\title{Tell Me Something New:\\ A New Framework for Asynchronous Parallel Learning}

\author{
  Julaiti Alafate \\
  University of California, San Diego \\
  \texttt{jalafate@eng.ucsd.edu} \\
  \And
  Yoav Freund \\
  University of California, San Diego \\
  \texttt{yfreund@ucsd.edu} \\
}

\begin{document}

\maketitle

\begin{abstract}
We present a novel approach for parallel computation in the context of
machine learning that we call ``Tell Me Something New'' (\tmsn). This
approach involves a set of independent workers that use broadcast to
update each other when they observe ``something new''. \tmsn\ does not
require synchronization or a head node and is highly resilient against
failing machines or laggards.
We demonstrate the utility \tmsn\ by applying it to learning boosted
trees. We show that our implementation is 10 times faster than
XGBoost~\cite{chen_xgboost:_2016} and LightGBM~\cite{ke_lightgbm:_2017} on the splice-site prediction
problem~\cite{sonnenburg_coffin:_2010, agarwal_reliable_2014}.
\end{abstract}

\section{Introduction}\label{sec:intro}

Ever-larger training sets call for ever faster learning algorithms.
On the other hand, computer clock rates are unlikely to increase
beyond 4\,GHz in the foreseeable future.  As a result there is a keen
interest in parallelized machine
learning algorithms~\cite{bekkerman_scaling_2012}.

The most common approach to parallel ML is based on Valiant's bulk
synchronous~\cite{valiant_bridging_1990} model. This approach calls for a
set of workers and a master. The system works in (bulk) iterations. In each iteration
the master sends a task to each worker and then waits for its
response. Once {\em all} machines responded, the master proceeds to the
next iteration. Thus the head node enforces synchronization (at the
iteration level) and maintains
a state that is shared by all of the workers.

Unfortunately, bulk synchronization does not scale well to more than
10--20 computers. Network congestion, latencies due to synchronization,
laggards, and failing computers result in diminishing benefits from
adding more workers to the
cluster~\cite{zaharia_apache_2016,mcsherry_scalability!_2015}.

There have been several attempts to break out of the bulk-synchronized
framework, most notably the work of Recht et~al.\ on
Hogwild~\cite{recht_hogwild:_2011} and Lian et~al.\ on asynchronous stochastic descent~\cite{lian_asynchronous_2015}. Hogwild
significantly reduces the synchronization penalty by using
asynchronous updates and parameter servers. The basic idea is to
decentralize the task of maintaining a global state and relying on
sparse updates to limit the frequency of update clashes.

\paragraph{Tell Me Something New}

Our first contribution is a new approach for parallelizing ML algorithms
which eliminates synchronization and the global state and instead uses a
distributed policy that guarantees progress. We call this approach
``Tell Me Something New'' (\tmsn). To explain \tmsn\ we start with an
analogy.

Consider a team of a hundred investigators that is going through
thousands of documents to build a criminal case where time is of the
issue. Assume also that most of the documents contain little or no new
information. How should the investigators communicate their findings
with each other? We contrast the bulk-synchronous (BS) approach and the
\tmsn\ approach. In the BS approach, each investigator takes a stack
of documents to their cubicle and reads through it. Then all of the
investigator meet in a room and tell each other what they found. Once
they are done, the process repeats. One problem with this approach is
that the fast readers have to wait for the slow readers. Another is
that a decision needs to be made as to how many documents or pages, to
put in each stack. Too many and the iterations would be very slow, too
few and all of the time would be spent in meetings.

The \tmsn\ approach is radically different. In this approach, each
investigator gets documents independently according to their speed of
reading and work habits. There is no meeting either. Instead, when
an investigator finds a piece of information that she believes is new,
she stands up in her cubicle and tells all of the other workers about
it. This has several advantages: nobody is ever waiting for anybody
else; the new information is broadcasted as soon as it is available, and
the system is fault resilient --- somebody falling asleep has little
effect on the others.
The analogy to parallel ML maps investigators
to computers, ``case'' to ``model'', and ``new information'' to
``improved model''.

More concretely, \tmsn\ for model learning works as follows. Each
worker has a model $H$ and an upper bound $L$ on the true loss of
$H$. The worker searches for a better model $H'$ whose loss upper
bound is $L'$. If $L'$ is significantly smaller than $L$, then the
worker takes two actions. First, $H',L'$ replaces $H,L$. Second
$(H',L')$ is broadcast to all other workers.  Each worker also listens
to the broadcast channel. If it receives pair $(H',L')$ it checks
whether $L'$ is significantly lower than its own upper bound $L$. If
it is, the worker replaces $(H,L)$ with $(H',L')$. Otherwise, the worker discards the
pair.

\paragraph{Boosting trees using TMSN}
Our second contribution is an application of \tmsn\ to boosted
decision trees. Boosted trees is a highly effective and widely used
machine learning method. In recent years there have been several
implementations of boosting that greatly improve over previous
implementations in terms of running time, in particular,
XGBoost~\cite{chen_xgboost:_2016} and
LightGBM~\cite{ke_lightgbm:_2017}. These implementations scale up to
training sets of many millions, or even billions of training examples.
Both implementations can run in one of two configurations: a
memory-only configuration where all of the training data is stored in
main memory, and a memory and disk configuration where the data is on
disk and is copied into memory when needed. The memory-only version is
significantly faster, but require a machine with very large
memory.

We present an implementation of boosting tree learning using
\tmsn\ that we call \Sparrow. This is a disk and memory
implementation, which requires only a fraction of the training data to
be stored in memory.  Yet, as our comparative experiments show, it is
about 10 times faster than XGBoost and LightGBM using the {\em memory
  only} configuration.

The rest of the paper is divided into four sections.
First we give a general description of \tmsn\ in Section~\ref{sec:tmsn}.
Then we introduce a special application of our algorithm, namely \Sparrow, in Section~\ref{sec:boost}.
After that we describe in more details of the algorithms and the system design of \Sparrow\ in
Section~\ref{sec:Algorithms}.
Finally, we present empirical results in Section~\ref{sec:experiments}.

\section{Tell Me Something New}\label{sec:tmsn}
We start with a general description of \tmsn\ which will be followed
by a description of \tmsn\ for boosting. To streamline our presentation
we consider binary classification, but other supervised or
unsupervised learning problem can be accommodated with little change.

We are given
\newcommand{\cD}{{\cal D}}
\begin{itemize}
\item A set of classifiers $\strongRules$, each classifier $H \in
  \strongRules$ is a mapping from an input space $X$ to a binary label $\{-1,+1\}$.
\item A stream of labeled examples $(x_1,y_1),(x_2,y_2),\ldots$, $x_i
  \in X$, $y_i \in \{-1,+1\}$, generated IID according to a fixed but
  unknown distribution $\cD$.
\end{itemize}

The goal of the algorithm is to find a classifier $H \in
\strongRules$ that minimized the error probability $\err(H)\doteq
P_{(x,y) \sim \cD}[H(x) \neq y]$

All workers start from the same initial classifier $H_0$ which is
improved iteratively. Some iterations end with the worker finding a
better classifier by itself, others end with the worker receiving a
better classifier from another worker. The sequences of classifiers
corresponding to different workers can be different, but with high
probability they all converge to the same classifier.

Denote each worker by an index $i=1,\ldots,n$. On iteration $t$
each worker has its {\em current} classifier  $H_i(t)$ and a set of $m$
{\em candidate} classifiers $G_i^j(t)$. An error upper bound
$\errub(H_i(t))$ is associated with $H_i(t)$ so that with high
probability $\errub(H_i(t)) \geq \err(H_i(t))$.

\begin{figure}[t]
\begin{center}
  \includegraphics[width=0.7\textwidth]{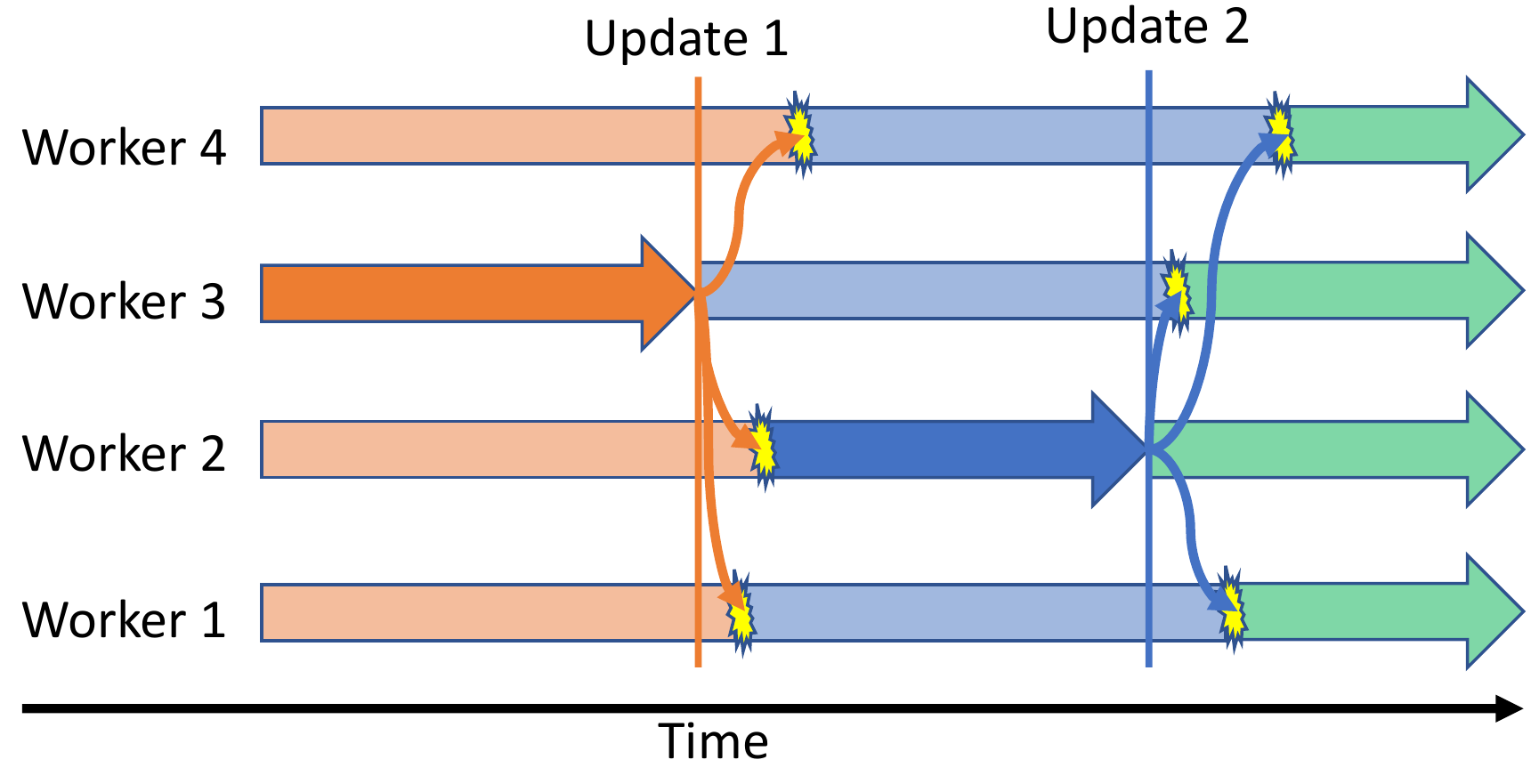}
\end{center}
  \caption{{\bf Execution timeline of a \tmsn\ system}
      System consists of four workers. The first update occurs when
      worker 3 identifies a better classifier $H_1$. It then replaces
      $H_0$ with $H_1$ and broadcasts $(H_1,z_1)$ to the
    other workers. The other workers receive the message the at different
    times, depending on network congestion. At that time they  interrupt the
    scanner (yellow explosions) and start using $H_1$. Next, worker 2
    identifies an improved rule $H_2$ and the same process ensues.
    \label{fig:async}}
   	\vspace{0pt}
\end{figure}

The worker reads examples from the stream and uses them to estimate
the errors of the candidates. It stops when it finds a candidate that,
with high probability, has an error smaller than
$\errub(H_i(t))-\epsilon$ for some constant ``gap'' parameter
$\epsilon>0$.

More precisely, the worker uses a {\em stopping rule} that chooses a
stopping time and a candidate rule and has the property that, with
high probability, the chosen candidate rule has an error smaller than
$\errub(H_i(t))-\epsilon$. This candidate then replaces the current
classifier, the new upper bound is set to be $\errub(H_i(t+1)) =
\errub(H_i(t))-\epsilon$, a new set of candidates is chosen and the
worker proceeds to the next iteration. At the same time the worker
{\em broadcasts} the pair $(H_i(t+1), \errub(H_i(t+1))$.

A separate process in each worker listens to broadcasts of this
type. When worker $i$ receives a pair $(H,\errub(H))$ it compares the
upper bound $\errub(H)$ with the upper bound associated with it's
current classifier $\errub(h_i(t))$. If $\errub(H) < \errub(h_i(t))-\epsilon$,
it interrupts the current search and sets $H_i(t+1)=H$. If not the
received pair is discarded.

Note that the only assumption that the workers make regarding the
incoming messages is that the upper bound $\errub(H)$ is sound. In
other words that, with hight probability, it is an upper bound on the
true error $\err(H)$. There is no synchronization and if a worker is
slow or fails, the effect on the other workers is minimal.

Different implementations of \tmsn\ differ in the way that they
generate candidate classifiers and in the stopping rules that they
use. For \tmsn\ to be effective, the stopping rule should be both
sound and tight. If it is not sound, then the scheme falls apart, and
if it is not tight, then the stopping rules stop later than needed,
slowing down convergence.

Next, we describe how \tmsn\ is applied to boosting.

\section{\tmsn\ for Boosting}~\label{sec:boost}
Boosting algorithms~\cite{schapire_boosting:_2012} are iterative, they generate a
sequence of {\em strong rules} of increasing accuracy. The strong rule
at iteration $T$ is a weighted majority over $T$ of the the weak rules
in $\cH$.
$$H_T(x) = \sign \left( \sum_{t=1}^T \alpha_t h_t(x) \right)$$

For the purpose of \tmsn\ we define $\strongRules$ to be the set of
strong rules combining any number of weak rules from $\cH$.

Boosting algorithms can be interpreted as gradient descent
algorithm~\cite{mason_boosting_1999}. Specifically, if we define the {\em potential} of
the strong rule $H$ with respect to the training set $\training$ to be
\[
\Z(H_T) \doteq \frac{1}{n} \sum_{i=1}^n e^{- y_i H_T(x_i)},
\]
then AdaBoost is equivalent to coordinate-wise gradient descent,
where the coordinates are the elements of $\cH$. Suppose we have the
strong rule $H$ and consider changing it to $H+\alpha h$ for some $h
\in \cH$ and for some small $\alpha$. The derivative of the potential
wrt $\alpha$ is:
$$
\left.\frac{\partial}{\partial \alpha}\right|_{\alpha=0} \Z(H+\alpha h) =
\frac{1}{n} \sum_{i=1}^n \left.\frac{\partial}{\partial \alpha}\right|_{\alpha=0} e^{ - y_i (H(x_i)+\alpha h(x_i))}
=
\frac{1}{n} \sum_{i=1}^n -y_i h(x_i) e^{- y_i H(x_i)}
$$
Our goal is to minimize the average potential $\Z(H_{T+1})$, therefor our goal is to
find a weak rule $h$ that makes the gradient negative. Another way of
expressing this goal is to find a weak rule with a large empirical {\em edge}:
\begin{equation} \label{eqn:gamma_emp}
\edgeEmp(h) \doteq  \sum_{i=1}^n w_i y_i h(x_i) \mbox{ where } w_i =
\frac{1}{Z}e^{- y_i H(x_i)}; Z = \sum_{i=1}^n e^{- y_i H(x_i)}
\end{equation}
$w_i$ defines a distribution over the training examples, with respect
to which we are measuring the correlation between $h(x_i)$ and $y_i$.
This is the original view of boosting, which is the
process of finding weak rules with significant edges with respect to
different distributions. We distinguish between the empirical
edge $\edgeEmp(h)$, which depends on the sample, and the {\bf true} edge, which
depends on the underlying distribution:
\begin{equation} \label{eqn:gamma_emp}
\edge(h) \doteq E_{(x,y) \sim \Dist}( w(x,y) y h(x)) \mbox{ where }
w(x,y)=\frac{1}{Z} \Dist(x,y) e^{- y H(x)}
\end{equation}
and $Z$ is the normalization factor with respect the the true
distribution $\Dist$.

A small but important observation is that boosting does not require
finding the weak rule with the {\bf largest} edge at each
iteration. Rather, it is enough to find a rule for which we are sure
that it has a significant (but not necessarily maximal)
advantage. More precisely, we want to know that, with high probability
over the choice of $\training \sim \Dist^n$  the rule $h$ has a significant
{\em true} edge $\edge(h)$.

\paragraph{Sequential Analysis and Early Stopping}\label{sec:methods:early-stop}
The standard approach when looking for the best weak rule
is to compute the error of candidate rules using
all available data, and then select the rule $h$ that maximizes the empirical
edge $\edgeEmp(h)$. However, as described above, this can be
over-kill. Observe that if the true edge $\edge(h)$ is large it can be
identified as such using a small number of examples.

Bradley and Schapire~\cite{bradley_filterboost:_2007} and Domingo and
Watanabe~\cite{domingo_scaling_2000} proposed using early stopping to
take advantage of such situations. The idea is simple: instead of
scanning through all of the training examples when searching for the
next weak rule, a {\em stopping rule} is checked for each $h \in \cH$
after each training example, and if this stopping rule ``fires'' then
the scan is terminated and the $h$ that caused the rule to fire is
added to the strong rule. We use early stopping in our algorithm.

For reasons that will be explained in the next section, we use a different
stopping rule than~\cite{bradley_filterboost:_2007}
or~\cite{domingo_scaling_2000}. We use a stopping rule proposed
in~\cite{balsubramani_sharp_2014} for which they prove the following

\begin{theorem}[based on \cite{balsubramani_sharp_2014} Theorem 4] \label{thm:balsubramani}
  Let $M_t$ be a martingale $M_t = \sum_i^t X_i$,
  and suppose there are constants $\{c_k\}_{k \geq 1}$ such that
  for all $i \geq 1$, $|X_i| \leq c_i$ w.p.\ 1.
  For $\forall \sigma > 0$, with probability at least $1 - \sigma$ we have
  \[
  \forall t: |M_t| \leq C \sqrt{
    \left( \sum_{i=1}^t c_i^2 \right)
    \left( \log \log \left( \frac{ \sum_{i=1}^t c_i^2 }{ |M_t| }\right) +
    \log \frac{1}{\sigma} \right)
  },
  \]
  where $C$ is a universal constant.
\end{theorem}

\paragraph{Effective Sample Size}
\label{sec:effectiveSampleSize}
Equation~\ref{eqn:gamma_emp} defines $\edgeEmp(h)$, which is an
estimate of $\edge(h)$. How accurate is this estimate? Our initial
gut reaction is that if $\training$ contains $n$ examples the error should be
about $1/\sqrt{n}$. However, when the examples are weighted this is
clearly wrong. Suppose, for example that $k$ out of the $n$ examples
have weight one and the rest have weight zero. Obviously in this case
we cannot hope for an error smaller than $1/\sqrt{k}$.

A more quantitative analysis follows. Suppose that the weights of the
examples in the training set $\training=\{ (x_1, y_1), \ldots, (x_n,
y_n) \}$ are $w_1=w(x_1,y_1),\ldots,w_n=w(x_n,y_n)$. Thinking of
finding a good weak rule in terms of hypothesis testing, the null
hypothesis is that the weak rule $h$ has no edge. Finding a rule that
is significantly better than random corresponds to rejecting the
hypothesis that $\edge(h)=0$.  Assuming the null hypothesis, $y_i
h(x_i)$ is $+1$ with probability 1/2 and $-1$ with probability
$1/2$. From central limit theorem and assuming $n$ is larger than
$100$, we get that the null distribution for $\edgeEmp(h)=\sum_{i=1}^n
w_i y_i h(x_i)$ is normal with zero mean and standard deviation
$\sum_{i=1}^n w_i^2$. The statistical test one would use in this case
is the $\Ztest$-test for
\begin{equation} \label{eqn:Ztest}
\Ztest = \frac{\edgeEmp(h)}{\sqrt{\sum_{i=1}^n w_i^2}}
= \frac{\sum_{i=1}^n w_i y_i h(x_i)}{\sqrt{\sum_{i=1}^n w_i^2}}
\end{equation}
As should be expected, the value of $\Ztest$ remains the same whether
or not $\sum_{i=1}^n w_i=1$. Based on Equation~\ref{eqn:Ztest} we
define the {\em effective number of examples} corresponding to the
un-normalized weights $w_1,\ldots,w_n$ as:
\begin{equation} \label{eqn:neff}
  \neff \doteq \frac{\left(\sum_{i=1}^n w_i\right)^2}{\sum_{i=1}^n w_i^2}
\end{equation}
Owen~\cite{owen_monte_2013} used a different line of
argument to arrived at a similar measure of
the effective samples size for a weighted sample.

The quantity $\neff$ plays a similar role in large deviation bounds
such as the Hoeffding bound~\cite{hoeffding_probability_1963} (details ommitted).
It also plays
a central role in Theorem~\ref{thm:balsubramani} and thus in the
stopping rule that we use.

To understand the important role that $\neff$ plays in our algorithm,
supppose the training set is of size $n$ and that only $m \ll n$
examples can fit in memory. Our approach is to start by placing a
random subset of size $m$ into memory and then run multiple
boosting iterations using this subset. As the strong rule improves,
$\neff$ decreases and as a result the stopping rule based on
Theorem~\ref{thm:balsubramani} requires increasingly more examples
before it is triggered. When $\neff/m$ crosses a pre-specified
threshold the algorithm flushes out the training examples currently in
memory and samples a new set of $m$ examples using acceptance
probability proportional to their weights. The new examples have
uniform weights and therefor after sampling $\neff=m$.

Intuitively, weighted sampling utilizes the computer's memory better
than uniform sampling because it places in memory more difficult
examples and fewer easy examples. The result is better estimates of
the edges of specialist\footnote{Specialist weak rules and their
  advantages are described in Section~{sec:Algorithm}} weak rules that
make predictions on high-weight difficult examples.

Another concern is the fraction of the examples that are selected. In
the method described here the expected fraction is $(\frac{1}{n}
\sum_{i=1}^n w_i)/(\max_i w_i)$.

\section{System architecture and Algorithms} \label{sec:Algorithms}

\begin{figure}
\centering
    \includegraphics[width=0.7\textwidth]{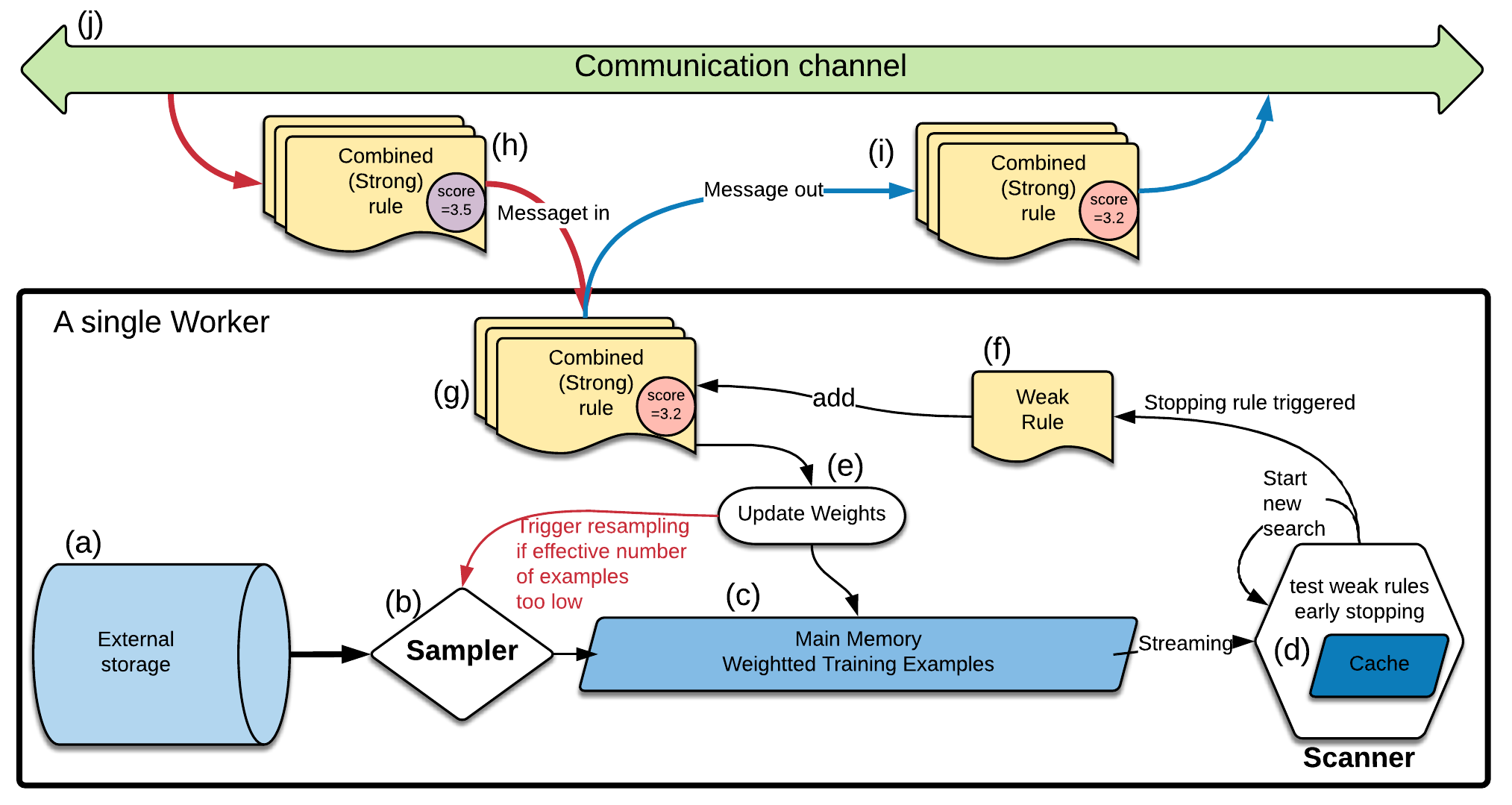}
    \caption{The \Sparrow\ system architecture.}\label{fig:architecture}
    \vspace{0pt}
\end{figure}

The \Sparrow\ distributed system consists of a collection of independent workers
connected through a shared communication channel. There is no
synchronization between the workers and no identified ``head node'' to
coordinate the workers. The result is a highly resilient system in which
there is no single point of failure and the overall slowdown resulting
from machine slowness or failure is proportional to the fraction of
faulty machines.

Each worker is responsible for a finite (small) set of weak
rules. This is a type of feature-based
parallelization~\cite{caragea_framework_2004}. The worker's task is to
identify a weak rule, based on one of the features in the set, that has
a significant edge.

We assume each worker stores {\em all} of the training examples
on it's local disk (element {\bf (a)} in Figure~\ref{fig:architecture})\footnote{In other
  words, the training data it replicated across all of the
  computers. This choice is made to maximize accuracy. If the data is
  too large to fit into the disk of a single worker, then it can be
  randomly partitioned between the computers. The cost is a potential increase
  in the difference between training error and test error}.

Our description of \Sparrow\ is in two parts. First, we describe the
design of a {\em single} \Sparrow\ worker. Following that, we describe
how concurrent workers use the \tmsn\ protocol to update each
other. Figure~\ref{fig:architecture} depicts the architecture of a
single computer and its interaction with the communication
channel. Pseudocode with additional detail is provided in the
supplementary material to this paper.

\subsection{A single \Sparrow\ worker}
\label{sec:single_worker}
As was said above, each worker is responsible for a set of the weak
rules.  The worker's task is to identify a rule that has a significant
edge (Equation~\ref{eqn:gamma_emp}). The worker consists of two
subroutines that can execute in parallel: a {\bf Scanner (d)} and a
{\bf Sampler (b)}. We describe each subroutine in turn.
\paragraph*{The Scanner}(element {\bf (d)} in
Figure~\ref{fig:architecture})
The Scanner's task is to read training examples sequentially and stop
when it has identified one of the rules to be a {\em good} rule. More
specifically, at any time point the Scanner stores the current strong
rule $H_t$, a set of candidate weak rules $\weakRules$ (which
define the candidate strong rules of \tmsn) and a target
edge $\gamma_t$. The scanner scans the training examples stored in
memory sequentially, one at a time. It computes the weight of the
examples using $H_t$ and then updates a running estimate of the edge
of each weak rule $h \in \weakRules$.

The scan stops when the stopping rule determine that
the true edge of a particular weak rule
$\gamma(h_t)$ is, with high probability,
larger than a threshold $\gamma$. The
worker then adds the identified weak rule $h_t$ {\bf (f)} to the current
strong rule $H_t$ to create a new strong rule $H_{t+1}$ {\bf (g)}.

The worker computes a ``performance score'' $z_{t+1}$ which is an
upper bound on the $Z$-score the strong rule by adding the weak rule
to it. The pair $(H_{t+1},z_{t+1})$ is broadcast to the other workers
{\bf (i)}. The worker then resumes it's search using the strong rule
$H_{t+1}$.
\paragraph*{The Sampler} Our assumption is that the entire training dataset does
not fit into main memory and is therefore stored in external storage
{\bf (a)}. As boosting progresses, the weights of the examples become
increasingly skewed, making the dataset in memory effectively smaller.
To counteract that skew, the {\bf Sampler} prepares a {\em new}
training set, in which all of the examples have equal weight, by using
selective sampling. When the effective number of examples associated
with the old training set becomes too small, the scanner stops using
the old training set and starts using the new one.\footnote{The
  sampler and scanner can run in parallel on separate cores. However in
  our current implementation the worker alternates between Scanning and
  sampling.}

The sampler uses selective sampling by which we mean that the
probability that an example $(x,y)$ is added to the sample is
proportional to $w(x,y)$. Each added example is assigned an initial
{\bf weight} of $1$.
\footnote{There are several known algorithms
  for selective sampling. The best known one is rejection sampling
  where a biased coin is flipped for each example. We use a method
  known as ``minimal variance sampling''~\cite{kitagawa_monte_1996}
  because it produces less variation in the sampled set.}
\paragraph*{Incremental Updates:} Our experience shows that the most
time consuming part of our algorithms is the computation of the
predictions of the strong rules $H_t$. A natural way to reduce this
computation is to perform it incrementally. In our case this is
slightly more complex than in XGBoost or LightGBM, because {\bf
  Scanner} scans only a  fraction of the examples at each
iteration. To implement incremental update we store for each example,
whether it is on disk or in memory, the results of the latest
update. Specifically, we store for each training example the tuple
$(x, y, w_s, w_l,H_l)$, Where $x,y$ are the feature vector and the
label, $H_l$ is the strong rule last used to calculate the weight of
the example. $w_l$ is the weight last calculated, and $w_s$ is
example's weight when it was last sampled by the sampler. In this way
{\bf Scanner} and {\bf Sampler} share the burden of computing
the weights, a cost that turns out to be the lion's share of the total
run time for our system.

\subsection{Communication between workers}
Communication between the workers is based on the \tmsn\ protocol.
As explained in Section~\ref{sec:single_worker}, when a worker
identifies a new strong rule, it broadcasts $(H_{t+1},z_{t+1})$ to all
of the other workers. Where $H_{t+1}$ is the new strong rule and
$z_{t+1}$ is an upper bound on the true $Z$-value of $H_{t+1}$. One
can think of $z_{t+1}$ as a ``certificate of quality'' for $H_{t+1}$.

When a worker receives a message of the form $(H,z)$,
it either accepts or rejects it. Suppose that the worker's current
strong rule is $H_t$ whose performance score is $z_t$. 
If  $z_t < z$ then the worker interrupts the Scanner and restarts it
with $(H_t,z_t) \gets (H,z)$.  If $z_t \geq z$ then $(H, z)$ is discarded
and the scanner continues running uninterrupted.

\section{Experiments}\label{sec:experiments}

In this section we describe the results of experiments comparing
the run time of \Sparrow\ with those of two leading implementations of
boosted trees: XGBoost and LightGBM.

\begin{table}[]
\centering
\label{table-exp}
\begin{tabular}{|l|l|c|c|}
\hline
Algorithm            & Instance & Instance Memory & Training (minutes)  \\ \hline
XGBoost, in-memory   & \texttt{x1e.xlarge}    & 122 GB          & 414.6                                  \\
XGBoost, off-memory  & \texttt{r3.xlarge}     & 30.5 GB         & 1566.1                                  \\
LightGBM, in-memory  & \texttt{x1e.xlarge}    & 122 GB          & 341.6                                  \\
LightGBM, off-memory & \texttt{r3.xlarge}     & 30.5 GB         & 449.7                                  \\
\multirow{2}{*}{TMSN, sample 10\%}    & \multirow{2}{*}{\texttt{c3.xlarge}}     & \multirow{2}{*}{7.5 GB}          & 57.4 (1 worker)      \\
                     &                        &                                    & 17.7 (10 workers) \\ \hline
\end{tabular}
\vspace{0.2cm}
\caption{Experiments on the Splice Site Detection Task}
\end{table}

\paragraph{Setup}
We use a large dataset that was used in other studies of large scale
learning on detecting human acceptor splice site~\cite{sonnenburg_coffin:_2010, agarwal_reliable_2014}.
The learning task is binary classification.
We use the same training dataset of 50\,M samples as in the other work,
and validate the model on the testing data set of 4.6\,M samples.
The training dataset on disk takes over 27\,GB in size.

As the code is not fully developed yet, we restrict our trees to one
level so-called ``decision stumps''. We plan to perform comparisons
using multi-level trees and more than two labels. We expect similar
runtime performance there. To generate comparable models,
we also train decision stumps in XGBoost and LightGBM
(by setting the maximum tree depth to 1).

Both XGBoost and LightGBM are highly optimized, and support multiple
tree construction algorithms.
For XGBoost, we selected approximate greedy algorithm for the efficiency purpose.
LightGBM supports using sampling in the training,
which they called \textit{Gradient-based One-Side Sampling (GOSS)}.
GOSS keeps a fixed percentage of examples with large gradients,
and then randomly sample from remaining examples with small gradients.
We selected GOSS as the tree construction algorithm for LightGBM.

All algorithms in comparison optimize the exponential loss as defined in AdaBoost.
We also evaluated the final model by calculating its area under precision-recall
curve (AUPRC) on the testing dataset.

Finally, the experiments are all conducted on EC2 instances from Amazon Web Services.
Since XGBoost requires 106\,GB memory space for training this dataset in memory,
we used instances with 120\,GB memory for such setting.
Detailed description of the setup is listed in Table~\ref{table-exp}.

\begin{figure}[t]
    \centering
    \includegraphics[width=0.65\textwidth]{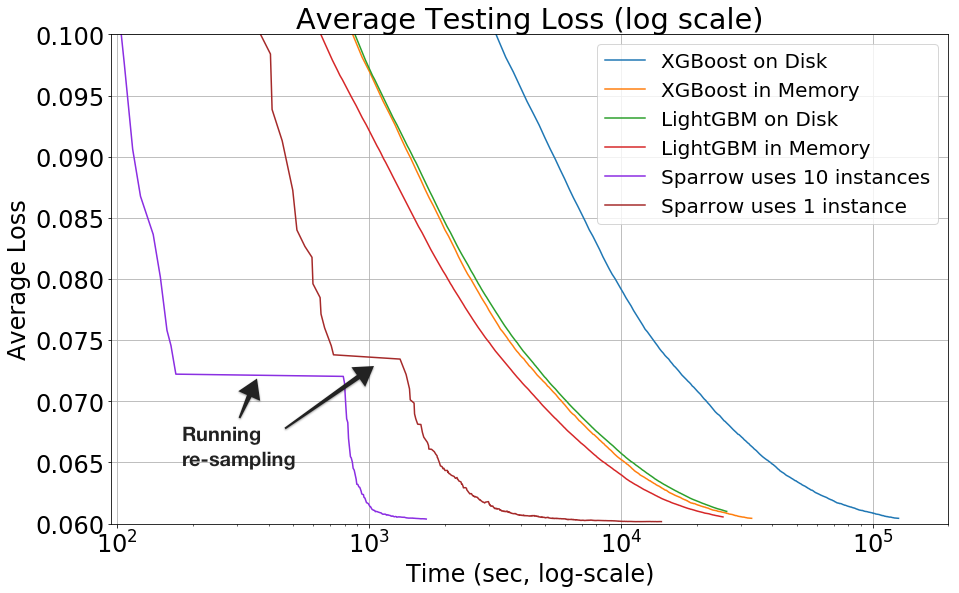}
    \caption{Comparing the average loss on the testing data using \Sparrow, XGBoost, and LightGBM, lower is better.
        The period of time that the loss is constant for \Sparrow\ is when the algorithm is generating a new sample set.}~\label{fig:loss}
\end{figure}

\paragraph{Evaluation}
Performance of each of the algorithm in terms of
the exponential loss as a function of time on the testing dataset is given in
Figure~\ref{fig:loss}. Observe that all algorithms achieve similar
final loss, but it takes them different amount of time to reach that
final loss. We summarize these differences in Table~\ref{table-exp} by
using the convergence time to an almost optimal loss of
$0.061$. Observe  XGBoost off-memory is about 27
times slower than a single \Sparrow\ worker which is also off-memory. That
time improves by another factor of 3.2 by using 10 machines instead of 1.

\begin{figure}[t]
    \centering
    \includegraphics[width=\textwidth]{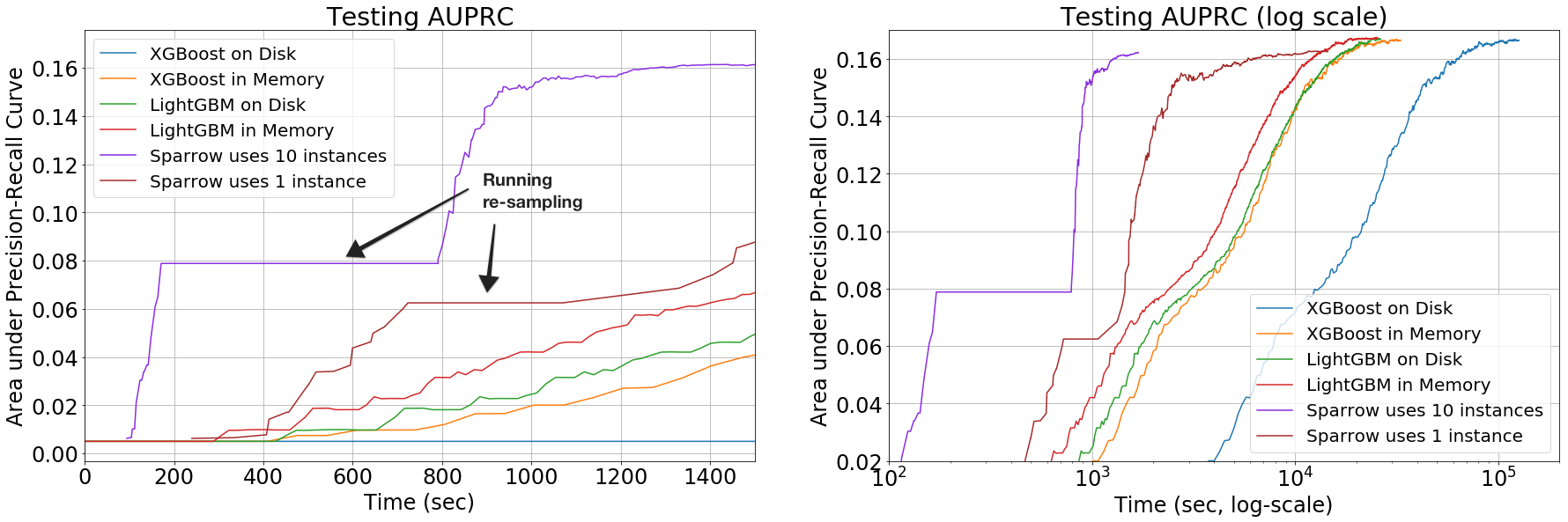}
    \caption{Comparing the area under the precision-recall curve (AUPRC) on the testing data
    using \Sparrow, XGBoost, and LightGBM, higher is better.
    (left) Normal scale, clipped on right.
    (right) Log scale, clipped on left.
    The period of time that the AUPRC is constant for \Sparrow\ is when the algorithm is generating a new sample set.}~\label{fig:auprc}
\end{figure}

In Figure~\ref{fig:auprc} we perform the comparison in terms of
AUPRC. The results are similar in terms of speed. However, in this
case XGBoost and LightGBM ultimately achieve a slightly better
AUPRC. This is baffling, because all algorithms work by minimizing
exponential loss.

\paragraph{Conclusions}
While the results are exciting plenty of work remains. We plan to
extend the algorithm to boosting full trees as well as other types of
classifiers.
In addition, we observe that run time is now dominated by the time it
takes to create new samples, we have some ideas for how to
significantly reduce the sampling time.

\clearpage
\small

\bibliographystyle{unsrt}
\bibliography{ms}

\clearpage
\section*{Appendix}

\algdef{SE}{When}{EndWhen}[1]{\textbf{when} \(\mbox{#1}\) \textbf{do}}{}%

\begin{algorithm}[H]
\caption{Procedures for the main algorithm and communication}\label{algorithm}

\begin{minipage}[t]{.48\textwidth}
\null

\begin{algorithmic}[0]

\State{\textbf{Procedure} \textit{MainAlgorithm}}
\State
\State \textbf{Initialize} $H=0, L=0$
\State \textbf{Create initial sample $S$} by calling \textsc{Sample}
\For{$k:=1 \ldots K$}
  \State $Ret \gets \Call{Scanner}{\gamma_0, M, i, H, \weakRules}$
  \If{$Ret$ {is} \textit{Fail}}
  \State Call \textsc{Sample} to Get a New Sample.
  \Else
  \State $i',h,\gamma \gets Ret$
  \State $i \gets i'$
  \State $H \gets H + \frac{1}{2} \log \frac{1/2+\gamma}{1/2-\gamma} h$
  \State Update $L$
  \EndIf
\EndFor

\end{algorithmic}


\end{minipage}%
\hspace{.04\textwidth}
\begin{minipage}[t]{.48\textwidth}
\null



\begin{algorithmic}[0]


\State{\textbf{Procedure} \textit{Communication}}
\State

\State \textit{(broadcasting out local models)}
\When{$H$ is updated in \textbf{this} worker}
\State \textbf{broadcast} $(H,L)$ to all other workers
\EndWhen

\State \textit{(receiving remote models from others)}
\When{received an $\left(H_{\mbox{new}},L_{\mbox{new}}\right)$ pair}
\If{$L_{\mbox{new}}<L$}
\State Interrupt \textsc{Scanner}
\State Replace $\left(H,L \right)$ with $\left(H_{\mbox{new}},L_{\mbox{new}}\right)$
\State Restart \textsc{Scanner}
\Else
\State Discard $\left(H_{\mbox{new}},L_{\mbox{new}} \right)$
\EndIf
\EndWhen

\end{algorithmic}
\end{minipage}

\end{algorithm}

\begin{algorithm}[H]

\caption{Functions for the Scanner and Sampler}\label{alg-scanner}

\begin{minipage}[t]{.47\textwidth}
\null
\begin{algorithmic}[0]

\State In-memory sampled set $S$ is defined globally
\State

\Function{Scanner}{$\gamma_0, M, i_0, H, \weakRules$}


\State $\gamma \gets \gamma_0, m \gets 0, i \gets i_0$
\State $V \gets 0, W \gets 0$
\State $\forall h \in \weakRules: m[h]=0$

\While {\textbf{True}}
\State $(x, y, w_s, w_l,H_l) \gets S[i]$  
\State $ i \gets (i+1) \mbox{ mod } |S| $
\If{$i=i_0$}
   \State \textbf{return} \textit{Fail}
\EndIf

\State $m \gets m + 1$
\If{$m>M$}
   \State $\gamma \gets \gamma/2$, $m \gets 0$
\EndIf
   
\State $w \gets \textsc{UpdateWeight}($
\State \hspace{1.0cm} $x,y,w_l,H_l,w_s,H,i)$  
\State $V \gets V+w^2$, $W \gets W + |w|$

\ForAll{$h  \in \weakRules$}
\State $m[h] \gets m[h] + w y h(x)$
\State $ret \gets \textsc{StoppingRule}($
\State \hspace{1.5cm} $W,V,m[h],\gamma)$

\If{$ret = \textbf{True}$}
\State \textbf{return} $i,h,\gamma$
\EndIf

\EndFor
\EndWhile

\EndFunction

\State

\end{algorithmic}

\end{minipage}
\begin{minipage}[t]{.52\textwidth}
\null

\begin{algorithmic}[0]

\Function{UpdateWeight}{$x,y,w_l,H_l,w_s,H,i$}
\State Calculate score update $s \gets H(x)-H_l(x)$
\State Calculate new weight $w \gets w_l \exp{(ys)}$ 
\State Update Sample: $S[i] = (x,y,w_s,w,H)$ 
\State \textbf{return} $w/w_s$
\EndFunction

\State

\Function{StoppingRule}{$W,V,m,\gamma$}
\State $C,\delta$ are global parameters.
\State $M \gets |m-2 \gamma W|$
\State \textbf{return} $M>C\sqrt{V(\log\log {V \over M_0}+ \log {1
    \over \delta}}$
\EndFunction

\State

\Function{Sample}{}

\State \textbf{Input:} Randomly permuted, disk-resident training-set.
\State \textbf{Input} \textrm{Current model} $H$

\State $S \gets \{\}$
\ForAll { \textrm{available training data } $(x, y)$ }
	\State $w_s \gets \exp{( -y H(x) )}$
	\State \textrm{With the probability proportional to } $w_s$, \\
	\hspace{1.5cm} $S \gets S + \{( x, y, w_s, w_s, H )\}$.
\EndFor

\State \textbf{return } S

\EndFunction

\end{algorithmic}

\end{minipage}
\end{algorithm}






\end{document}